\documentclass{article}

\usepackage{microtype}
\usepackage{graphicx}
\usepackage{subfigure}
\usepackage{booktabs} % for professional tables
\usepackage{amsmath, amssymb, amsthm} 
\usepackage{color}
\usepackage{algorithm2e}
\usepackage{graphicx}% http://ctan.org/pkg/graphicx
\graphicspath{ {./figures/} }
\usepackage{multirow}
\usepackage{url}
\newcommand{\x}{\mathbf{x}}
\newcommand{\y}{\mathbf{y}}
\newcommand{\z}{\mathbf{z}}
\newcommand{\uu}{\mathbf{u}}
\newcommand{\vv}{\mathbf{v}}
\newcommand{\w}{\mathbf{w}}
\newcommand{\dd}{\mathbf{d}}

\usepackage{hyperref}

\usepackage[accepted]{icml2021}

\icmltitlerunning{Integrating Black-Box Scientific Knowledge with Deep Learning}

\begin{document}

\twocolumn[
\icmltitle{\emph{Zeroth-Order SciML}: Non-intrusive Integration of Scientific Software with Deep Learning}




\begin{icmlauthorlist}
\icmlauthor{Ioannis Tsaknakis}{to}
\icmlauthor{Bhavya Kailkhura}{goo}
\icmlauthor{Sijia Liu}{ed}
\icmlauthor{Donald Loveland}{ar}
\icmlauthor{James Diffenderfer}{goo}
\icmlauthor{Anna Maria Hiszpanski}{goo}
\icmlauthor{Mingyi Hong}{to}
\end{icmlauthorlist}

\icmlaffiliation{to}{University of Minnesota}
\icmlaffiliation{goo}{Lawrence Livermore National Laboratory}
\icmlaffiliation{ed}{Michigan State University}
\icmlaffiliation{ar}{University of Michigan, Ann Arbor}

\icmlcorrespondingauthor{Bhavya Kailkhura}{kailkhura1@llnl.gov}

\icmlkeywords{Machine Learning, ICML}

\vskip 0.3in
]




\printAffiliationsAndNotice{} % otherwise use the standard text.

\begin{abstract}
Using deep learning (DL) to accelerate and/or improve scientific workflows can yield discoveries that are otherwise impossible. Unfortunately, DL models have yielded limited success in complex scientific domains due to large data requirements. In this work, we propose to overcome this issue by integrating the abundance of scientific knowledge sources (SKS) with the DL training process. Existing knowledge integration approaches are limited to using differentiable knowledge source to be compatible with first-order DL training paradigm. In contrast, our proposed approach treats knowledge source as a black-box in turn allowing to integrate virtually any knowledge source. To enable an end-to-end training of SKS-coupled-DL, we propose to use zeroth-order optimization (ZOO) based gradient-free training schemes, which is non-intrusive, i.e., does not require making any changes to the SKS. 
%Unlike the current gradient-based training paradigm, proposed techniques are able to train SKS-coupled-DL models accurately using only function queries, thereby bypassing the stringent requirement of gradient availability. 
We evaluate the performance of our ZOO training scheme on two real-world material science applications. 
We show that proposed scheme is able to effectively integrate scientific knowledge with DL training and is able to outperform purely data-driven model for data-limited scientific applications. 
We also discuss some limitations of the proposed method and mention potentially worthwhile future directions. 
\end{abstract}

\section{Introduction}
Deep learning (DL) provides incredible opportunities to answer some of the most important and difficult questions in a wide range of scientific applications. 
However, many of the scientific use cases such as materials discovery do not have access to large amounts of data required to train DL models.   
In such data-limited complex scientific domains, purely data-driven DL models have yielded limited success or led to unsatisfactory results.
At first glance, the task of training an accurate deep neural network (DNN) from a small amount of data appears impossible. 
Coming to the rescue is a wealth of prior domain knowledge in the form of scientific software codes -- most of which is currently not being utilized as purely-data driven models inefficiently try to learn already known chemistry from scratch.

The biggest challenge in integrating scientific knowledge into the learning process is formulating it in a manner that is compatible with the current training paradigm. 
Given the fact that nearly all DNN training approaches are based on differentiable programming, all existing efforts in integrating scientific knowledge\cite{von2019informed} are either (a) restricted to using a very limited subset of knowledge that is already in a differentiable form, or (b) spending significant effort on modifying knowledge sources to support automatic differentiability (autodiff).
As an example of (a), Physics-aware machine learning (ML) area~\cite{karniadakis2021physics} uses analytical equations, which only capture a very limited aspect of otherwise complex system. 
As an example of (b), Differentiable Physics/Chemistry area~\cite{diff1, diff2} requires updating, or altogether rewriting, existing software codes to provide autodiff functionality, which is extremely costly. 
This points to a crucial roadblock in using DL in scientific domains: we do not have methods that can exploit all available scientific knowledge without being intrusive, i.e., requiring a rewrite of existing software codes. 

\begin{figure*}[t]
    \centering
    \includegraphics[scale=0.45]{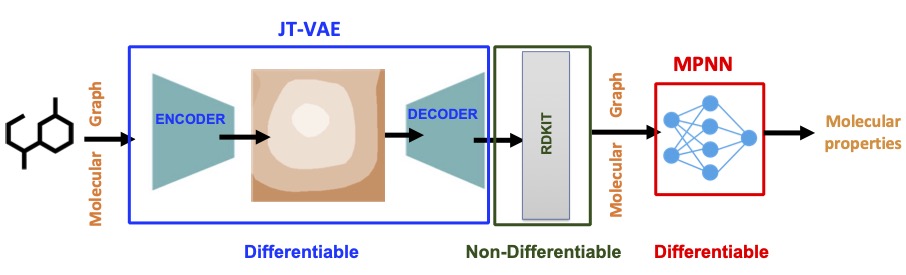}
    \caption{Scientific machine learning pipeline for molecular property prediction and generation.}
    \label{fig:pipeline}
\end{figure*}

To overcome the challenge of effortless integration of scientific knowledge with DL, we propose making changes to the ML paradigm to be compatible with scientific software, rather than the other way around. %like current trend. 
Specifically, we propose to leverage gradient-free optimization approaches to train scientific software coupled deep neural nets. 
Note that resulting ``sciML" system does not provide gradients rendering popular first-order methods (e.g., stochastic gradient descent) inapplicable. 
Unlike the current gradient-based training paradigm, i.e., backprop, our proposed technique aims to train sciML models accurately using only function queries (or forward passes), thereby bypassing the stringent requirement of gradient availability. 
To achieve this, we adopt a subset of gradient-free optimization, Zeroth-Order Optimization (ZOO)~\cite{liu2020primer, liu2018zeroth}, and train sciML system in an end-to-end manner. 
To show the feasibility of our approach, we apply our technique on a popular material discovery pipeline on two datasets. This pipeline is used to produce novel molecules with desired properties by searching for molecular graphs in the graph generative model's latent space. The sciML system for this problem is comprised of the following three components: 1) Generative model: junction tree variational autoencoder (JT-VAE), 2) Molecular Featurizer: RDKit cheminformatics software, and 3) Predictive model: message passing neural network (MPNN) (see Figure~\ref{fig:pipeline}). 
Note that the reasoning behind using a generative model is to transform the original discrete space where molecules lie to a continuous space that is amenable for search and optimization.  
Unfortunately, the presence of RDKit software (which is effectively a black-box function) renders the sciML system non-differentiable. 
This precludes the use of gradient-based methods for end-to-end training. 
On the other hand, existing approaches resort to poor workaround of using purely data-driven models. 
These approaches pretrain the JT-VAE and MPNN separately using first-order methods. 
However, in this approach, the JT-VAE training if not aware of the property of interest predicted by MPNN. Thus, JT-VAE latent space often is not amenable to maximizing the property predicted by MPNN. More importantly purely data-driven approaches require a large amount of data for learning generalizable patterns. 
%\textcolor{blue}{I think this could use further motivation, what could go wrong if they arent amenable? I.e., falling out of support of the generative model dist may lead to [blank]}
%{\color{red} Justify why this pipeline. JT-VAE for continuous latent space for search.}
%Another popular approach is to directly fit a Gaussian Process (GP) model in the latent space (avoiding reconstruction of the molecular graph). This approach suffers from maintaining the molecular validity of the solution. 

%\textcolor{blue}{May be good to do a bit of rephrasing here, the paragraph seems to jump back and forth between past and present tense. I also see phrasing that uses "we aim" and "we believe", which maybe can be dropped since it either did or didnt happen. This is also sounds like the core contributions paragraph of the intro which may be worth highlighting.}
In this paper, on the other hand, we aim to perform an end-to-end training of sciML sysyem in Figure~\ref{fig:pipeline}. 
With a clever application of zeroth-order methods, we manage to overcome this obstacle and develop gradient-free end-to-end training algorithms. We show that the proposed zeroth-order training approach can outperform purely data-driven first-order methods by exploiting the knowledge encoded in the scientific software RDKit. This is the key contribution of this paper, and we hope that its application extends beyond the material science use case considered in this work.

\section{Problem Formulation}
Although our formulation of integrating black-box knowledge with DNN is quite general, we concretize our discussion with a specific real-world material science application.

\subsection{Molecular Property Prediction System}
Here we provide details on our molecular property prediction system (Figure~\ref{fig:pipeline}) that serves as a fundamental building block for molecular screening and discovery problems.

Let us define a dataset $D = \{(\x_{i}, \y_{i})\}_{i=1}^{N}$ of $N$ molecules, where $\x_i$ is the graph $G=(V,E)$ of a single molecule
%\footnote{The datasets we are using in this work represent molecules as SMILES strings \cite{weininger1988smiles}. However, the main models we are using in this  work apply directly on molecular graphs. In addition, transforming a SMILES string into a suitable graph representation is easy, and thus we can assume that the dataset ``feature vectors'' are graphs.} 
with vertex features $\{\mathbf{s}_u\}_{u \in V}$ and edge features $\{\mathbf{e}_{uv}\}_{uv \in E}$; the vertex and edge features describe certain atom and bond characteristics, respectively. The vector $\y_i$ denotes certain chemical properties, such as density and heat of formation.

The predictive system of interest consists of the following three distinct components.
\begin{enumerate}
    \item \underline{Generative model} (JT-VAE)
    
    We use the Junction Tree Variational Auto-Encoder (JT-VAE) \cite{jin2018junction} tailored for working on molecular graphs. It consists of: 1) an encoder which is a neural network $F_{E}(\cdot;\uu)$ that maps the input $\x$ (molecular graph) to a latent representation $\z$ where $\uu$ denotes the encoder parameters, and 2) a decoder which is a neural network $F_{D}(\cdot;\vv)$ that maps (reconstructs) the latent representation $\z$ back to a molecular graph $\tilde{\x}$ where $\vv$ denotes the decoder parameters.
    % The training of the JT-VAE aims to learn a continuous latent space from which we can sample molecules.
    
    \item \underline{Molecular Featurizer} (RDKit)
    
    JT-VAE output is followed by a cheminformatics software that takes a molecule (in the SMILES string form) as input, computes certain molecular descriptors, and outputs a molecular graph where these descriptors are then incorporated as node and edge features. We use the RDKit library \cite{Landrum2016RDKit} as our cheminformatics knowledge base. For each problem, any feature that has a constant value across the whole dataset is removed prior to training. 
    From a modelling perspective we treat this component as a black box $G(\cdot)$ in which we can send queries (molecules) $\tilde{\x}$ and get molecular graphs $\hat{\x}$ with certain edge and node features as outputs. However, note that the black-box nature of $G$ prevents us from having direct access to its gradients, and  thus the pipeline is effectively rendered non-differentiable.
    
    \item \underline{Predictive model} (MPNN)

    We use the Message Passing Neural Net (MPNN) as our predictive model, which we denote with $H(\cdot;\w)$ where $\w$ denotes the model parameters. MPNN takes a molecular graph $\hat{\x}$ as an input and predicts certain molecular properties. Note that MPNN is a graph neural network (GNN); thus, it offers the advantage of leveraging node/edge features and connectivity available in the network structure of a molecule.
\end{enumerate}

Using the above notations, the predicted property of molecule $i$ is denoted as
\begin{align}\label{eq:model}
    \widetilde{\y}_{i} = H\left( G\left( F_{D}\left( F_{E}\left(\x_{i};\uu \right);\vv \right) \right);\w \right).
\end{align}

\subsection{RDKit Integrated Training of JT-VAE and MPNN}

%\textcolor{blue}{I am not sure it is totally clear to me why being able to jointly train would improve predictive performance. I think this is largely because the motivation behind why training JTVAE and MPNN separately could be problematic isnt clear enough.}

Note that material science applications usually are data-limited, which makes it difficult for purely data-driven models to learn generalizable patterns.  
In this work, we aim to improve the performance of the aforementioned prediction system by leveraging end-to-end training with RDKit as an additional knowledge source. 
The objective of the training procedure is to jointly optimize the parameters of JT-VAE (i.e., encoder and decoder) and MPNN. This goal is achieved by solving the following optimization problem:
\begin{align}
\label{eq:opt}
    &\min_{\uu,\vv,\w} L(\x, \y;\uu,\vv,\w),\\
    &L = 
     L_{mpnn}(\x, \y;\uu,\vv,\w) + \lambda\; L_{jtvae} \left (\x;\uu, \vv\right), \nonumber\\
    & L_{mpnn} = 
    \frac{1}{N} \sum\limits_{i=1}^{N} \|\widetilde{\y}_{i}-\y_i \|^{2}, \nonumber\\
    & L_{jtvae} =  
    \ell \left (F_{D}\left( F_{E}\left(\{\x_i\}_{i=1}^{N};\uu \right);\vv \right), \{\x_i\}_{i=1}^{N} \right),\nonumber
\end{align}
where $\ell(\cdot,\cdot)$ denotes the JT-VAE loss (for more details refer to \cite{jin2018junction}), $\lambda$ is a regularization parameter, and $\widetilde{\y}_{i}$ is defined in \eqref{eq:model}. Note that $L_{mpnn}$ is a function of RDKit output, which makes it necessary to use gradient-free optimization methods.

\section{Proposed Zeroth-Order Training Algorithm}
To solve \eqref{eq:opt}, we adopt a variant of gradient-free optimization, Zeroth-Order Optimization (ZOO), which operates similarly to iterative methods: gradient estimation (via function evaluations), descent direction computation, and solution update. Thus, our ZOO end-to-end training replaces gradients with their estimates (whose exact form is defined in the next section) and proceeds as traditional training.

\subsection{ZOO for the sciML System in Figure~\ref{fig:pipeline}}
Specifically, the gradient of loss $L$ (w.r.t. to the model parameters $\uu,\vv,\w$) are as follows: 
\begin{align*}
    & \nabla_{\w} L = \nabla_{\w} L_{mpnn}(\{(\x_i, \y_i)\}_{i=1}^{N};\uu,\vv,\w)\\
    & \widetilde{\nabla}_{\uu,\vv} L =
    \widetilde{\nabla}_{\uu,\vv} L_{mpnn}(\{(\x_i, \y_i)\}_{i=1}^{N};\uu,\vv,\w) \\&\quad\quad\quad+  \lambda \nabla_{\uu,\vv} L_{jtvae} \left (\{\x_i\}_{i=1}^{N};\uu, \vv\right) 
\end{align*}    
\begin{align*}    
    &= \frac{1}{N} \sum\limits_{i=1}^{N} 2 \left( \widetilde{\y}_i-\y_i \right) \widetilde{\nabla}_{\uu,\vv} \widetilde{\y}_i +  \lambda \nabla_{\uu,\vv} L_{jtvae},
\end{align*}
where the symbol $\widetilde{\nabla}$ is used to denote a gradient estimate.

Using the chain rule we have
\begin{small}
\begin{align}
    &\widetilde{\nabla}_{\uu} \widetilde{\y}_i
    = \nabla_{\uu} F_{E}(\x_i;\uu) 
    \nabla_{1} F_{D}\left( F_{E}\left(\x_i;\uu \right);\vv \right)\nonumber\times\\ &
    \widetilde{\nabla}  G\left( F_{D}\left( F_{E}\left(\x_i;\uu \right);\vv \right) \right) 
    \nabla_{1} H\left( G\left( F_{D}\left( F_{E}\left(\x_i;\uu \right);\vv \right) \right);\w \right) \label{eq:chain1}\\
    &\widetilde{\nabla}_{\vv} \widetilde{\y}_i  = \nabla_{\vv} F_{D}\left( F_{E}\left(\x_i;\uu \right);\vv \right)
    \widetilde{\nabla}  G\left( F_{D}\left( F_{E}\left(\x_i;\uu \right);\vv \right) \right)\nonumber\\ &\quad\quad\quad\times 
    \nabla_{1} H\left( G\left( F_{D}\left( F_{E}\left(\x_i;\uu \right);\vv \right) \right);\w \right),\label{eq:chain2}
\end{align}
\end{small}%
where $\nabla_{1}$ denotes the gradient w.r.t. the first argument (i.e., the model's input) and $\widetilde{\nabla} G$ some gradient estimate of the black-box function $G(\cdot)$.

\paragraph{Challenges with sciML System in Figure~\ref{fig:pipeline}.}
Upon a closer examination of aforementioned ZOO steps, notice some major technical hurdles arise due to subtle application level details. Next, we explain these  issues in detail.   
\begin{enumerate}
    \item The gradients involved in \eqref{eq:chain1}, \eqref{eq:chain2} are not well-defined for the application of interest. Specifically, notice that 
    \begin{itemize}
        \item The term
         $\widetilde{\nabla}  G\left( F_{D}\left( F_{E}\left(\x;\uu \right);\vv \right) \right)$ denotes the gradient estimate of graph w.r.t. SMILES string.
        \item The term
         $\nabla_{1} H\left( G\left( F_{D}\left( F_{E}\left(\x;\uu \right);\vv \right) \right);\w \right)$ denotes the gradient of a scalar w.r.t. a graph.
         
    \end{itemize}
    \item The gradient estimates we aim to form for black-box function $G(\cdot)$ will involve a quantity of the form 
    $
        G(\x + \mu \mathbf d) - G(\x)
    $, (which typically appears in zeroth-order gradient estimators)
    where $\x$ is a molecule and $\mathbf d$ is a direction in the molecules space (the exact representation, e.g., SMILES string or graph, is not relevant at this point); the output of $G(\cdot)$ is a graph. Then, it is not clear what the meaning of the above difference is.
\end{enumerate}

\begin{figure*}[t]
    \centering
    \includegraphics[scale=0.45]{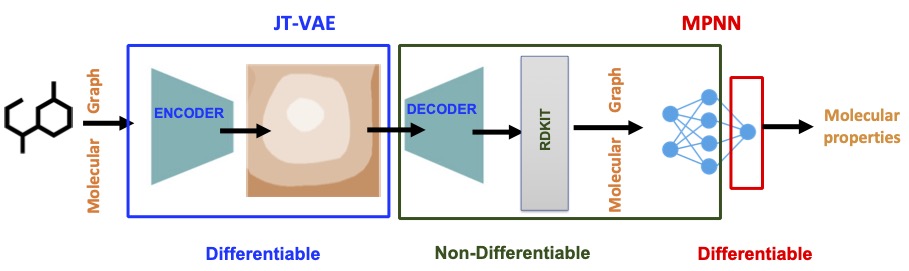}
    \caption{We change the boundaries of the three components for our algorithm design compared to the original one in Fig. \ref{fig:pipeline}. }
    \label{fig:pipeline1}
\end{figure*}

\subsection{Reparameterization to the Rescue}
From the above discussion, it is clear that we need to use reparameterization in order to resolve these issues. 
Our approach is to adjust the boundaries of the three components that comprise the system (see Figure~\ref{fig:pipeline1}). Specifically, we perform the following changes:
\begin{itemize}
    \item The boundary between the 1st component and the 2nd component becomes the intermediate latent layer of the JT-VAE. We maintain the same notation as previously and denote the input-output relation with $F_{E}\left(\x;\uu \right)$.
    
    \item The 2nd component includes the layers of the MPNN which performs the embedding of the graph into a continuous latent space. Specifically, the output of the 2nd component is the readout function of the MPNN. Note that the presence of the RDKit module in this component renders the pipeline incompatible to first-order methods. We denote this component with $\widetilde{G}(\cdot; \vv, \w_{1})$, where $\w_{1}$ includes the  parameters of the MPNN inside this component.
    
    \item The third component now includes only the part of the MPNN after the readout function. We denote the parameters involved in this part with $\w_{2}$ and the input-output relation as $\widetilde{H}(\cdot;\w_{2})$.
\end{itemize}

The output of the reparameterized system can be written as
$\widetilde{\y} = \widetilde{H}(\widetilde{G}(F_{E}(\x_i,\uu); \vv, \w_{1}); \w_{2})$.
As a result, the chain rule is rearranged as:
\vspace{-0.1in}
\begin{align*}
    &\widetilde{\nabla}_{\uu} \widetilde{\y}_{i}
    = \nabla_{\uu} F_{E}(\x_i;\uu) \widetilde{\nabla}_{1} \widetilde{G}(F_{E}(\x_i;\uu);\vv,\w_{1}) \\&\quad\quad\quad \times \nabla_{1} \widetilde{H}(\widetilde{G}(F_{E}(\x_i;\uu);\vv,\w_{1}); \w_{2})\\
    &\widetilde{\nabla}_{\vv} \widetilde{\y}_{i}
    = \widetilde{\nabla}_{\vv} \widetilde{G}(F_{E}(\x_i;\uu);\vv,\w_{1}) \\&\quad\quad\quad \times \nabla_{1} \widetilde{H}(\widetilde{G}(F_{E}(\x_i;\uu);\vv,\w_{1}); \w_{2}).
\end{align*}

\noindent \textbf{Zeroth-order estimators.}
One aspect we have not addressed so far is the specific choice for the gradient estimator of the black-box function $\widetilde{\nabla}_{1} \widetilde{G}(\cdot), \widetilde{\nabla}_{\vv} \widetilde{G}(\cdot)$. In this work, we employ a coordinate-wise zeroth-order (ZO) gradient estimator~\cite{liu2020primer}:

\begin{align*}
& \widetilde{\nabla}_{1} \widetilde{G}(F_{E}(\x;\uu);\vv,\w_{1}) =\\
&\frac{\widetilde{G}(F_{E}(\x;\uu) + \mu_1 \dd_1;\vv,\w_{1}) - \widetilde{G}(F_{E}(\x;\uu);\vv,\w_{1})}{\mu_1} \dd_1\\
& \widetilde{\nabla}_{\vv} \widetilde{G}(F_{E}(\x;\uu);\vv,\w_{1})=\\
&\frac{\widetilde{G}(F_{E}(\x;\uu);\vv+ \mu_2 \dd_2,\w_{1}) - \widetilde{G}(F_{E}(\x;\uu);\vv,\w_{1})}{\mu_2} \dd_2,
\end{align*}
where $\mu_{1},\mu_{2}$ are the smoothing parameters, and $\dd_{1},\dd_{2}$ are directions in the (continuous) latent space; the elements of these direction vectors are Gaussian random variables of mean $0$ and variance $\sigma^{2}$.

\section{Experiments}

\paragraph{Goal:} Our experiments try to find: can the ZO training approach  integrate black-box knowledge source in sciML learning process to learn more generalizable models despite of data scarcity?

\paragraph{Datasets:}
We consider the following two datasets:
\begin{enumerate}
    \item ZINC \cite{zinc}: We have $5500$ molecules in the training set and $5000$ molecules in the test set. Similar to \cite{kusner2017grammar,jin2018junction}, the property we predict is a penalized version of the water-octanol partition coefficients ($\log P$), specifically the quantity $\log P-SA-RS$, where $SA$ is the synthetic accessibility score and $RS$ is the ring size.
    
    \item HOF~\cite{griessen1988heat} dataset: This dataset contains $5600$ molecules in the training set and $7000$ in the test set. The property of interest is the heat of formation.
\end{enumerate}

\paragraph{Models and Hyperparameters:}

For each dataset we perform the following set of experiments:
\begin{enumerate}
    \item \underline{Stage1}: We train a JT-VAE and a MPNN model independently (or in a purely data-driven manner) using the DGL-LifeSci package\footnote{\url{https://github.com/awslabs/dgl-lifesci}}.

    \item \underline{Stage2}: For Stage2 models, we inject RDKit knowledge into the purely data-driven models by using zeroth-order end-to-end training.  
\end{enumerate}

Below we provide details on our hyperparameter selection, which remain the same across both experiments.
    \begin{enumerate}
        \item \underline{JT-VAE}: We use learning rate $=0.0007$, network depth$=3$, size of hidden layers$=450$ (for both the encoder and the decoder), and dimension of latent space$=56$.
        \item \underline{MPNN}: We use learning rate$=0.0005$, size of node input features$=39$, size of edge input features$=50$, size of output node representations$=128$, size of hidden edge representations$=64$, and number of message passing steps$=6$.
    \end{enumerate}
Note that the batch size was $16$ for Stage1 experiments, and $8$ for Stage2. Moreover, in Stage2 experiments we employ gradient clipping in order to deal with exploding gradients. 
We also set $\lambda=1$ and freeze the parameters of the decoder (this is a reasonable choice as the decoder was already pretrained sufficiently well in Stage1). Finally, zeroth-order gradient estimators uses $\mu_{1}=\mu_{2} = 0.001$.  

\paragraph{Metrics:}
We compare the performance of our approach w.r.t. baselines in terms of following metrics:
\begin{enumerate}
    \item Reconstruction ability of the generative model.
    
    The molecule is passed through the JT-VAE, where it is first decoded (i.e., we obtain its continuous representation in the latent space), and then encoded back into a (graph) molecular form. We consider the reconstruction successful if the molecule we get in the output (after the decoder) is the same as the molecule we provide as input (before the encoder). We report the reconstruction accuracy, which is defined as $$\text{accuracy} = \frac{\# \text{successfully reconstructed molecules}}{\text{total number of molecules evaluated}}.$$
    
    \item Predictive ability of the supervised (predictive) model
    
    We compute the Root Mean Square Error (RMSE) over the dataset, between the true value of the molecular property and the value predicted by the MPNN,
    $$\text{RMSE}= \left[ \frac{1}{N} \sum\limits_{i=1}^{N} \left( y_{i} - \widetilde{y}_{i} \right)^{2} \right]^{1/2}.$$
\end{enumerate}

\paragraph{Results:}
Table \ref{tab:results} presents the results of the experiments described above.
Note that MPNN is trained in Stage 1 for a maximum number of epochs such that we can no longer improve its performance (i.e., decrease RMSE). Then, notice that the jointly training the MPNN (along with the JT-VAE) leads to better prediction performance, under both datasets. Although we observe a decrease in the reconstruction error, note that reconstruction is not our main objective. In fact the main motivation behind using JT-VAE is to obtain a continuous latent space. Regardless, a successful training by simply using forward passes (i.e., without gradients) is exciting and motivates more research in this direction.

\begin{table}[]
\centering
\begin{tabular}{|c|c|c|c|}
\hline
Dataset                        & Model (Measure)                           & Stage1       & Stage2 \\ \hline
\multirow{2}{*}{\textbf{ZINC}} & \textit{JT-VAE (Recon. accuracy)} & $4.59\%$   & $3.97\%$     \\ \cline{2-4} 
                               & \textit{MPNN (RMSE) with JT-VAE}                      & $1.83$     & $\textbf{1.66}$    \\ \hline
\multirow{2}{*}{\textbf{HOF}}  & \textit{JT-VAE (Recon. accuracy)} & $22.5\%$   & $15.5\%$     \\ \cline{2-4} 
                               & \textit{MPNN (RMSE) with JT-VAE}                      &  $53$      &  $\textbf{52.2}$    \\ \hline
\end{tabular}
\caption{Performance comparison of purely data-driven models (trained with first-order method) with RDKit integrated models (trained with proposed zeroth-order method).}
\label{tab:results}
\end{table}

We also mention some limitation of the proposed ZOO scheme. 
First, ZOO requires \textit{large runtime} mainly due to application related factors that we cannot completely control. 
For instance, the decoding phase of the JT-VAE, which is involved in the computation of $\widetilde{G}(F_{E}(\x;\uu) + \mu_1 \dd_1;\vv,\w_{1})$ and $\widetilde{G}(F_{E}(\x;\uu);\vv+ \mu_2 \dd_2,\w_{1})$ (in the ZO estimates) can be very slow, especially for novel molecules. Second, JT-VAE requires pretraining in order to decode molecules reasonably fast. Finally, training requires carefully balancing prediction error vs. reconstruction accuracy. Choosing right hyperparameters is critical to a stable training.

\section{Conclusion}
In this paper, we developed gradient-free training approaches for enabling integration of black-box scientific knowledge sources with deep learning. 
Using two materials datasets, we showed that proposed knowledge-integrated models outperform purely data-driven models in small-data regime. In the future we will explore the use of more sophisticated zeroth-order gradient estimators, e.g., average several gradient estimates across different directions. Another worthwhile direction is to apply the proposed approach to additional applications where scientific knowledge is available in the form of a black-box. Our results provide a new perspective on the integration of knowledge sources with deep learning. We hope that our positive results   
will encourage deep learning community to pay more attention to ZOO-based knowledge integration. 

\section*{Acknowledgement}
This work was performed under the auspices of the U.S. Department of Energy by Lawrence Livermore National Laboratory under Contract DE-AC52-07NA27344. James Diffenderfer's effort was supported by the LLNL-LDRD Program under Project No. 22-FS-019.

\bibliographystyle{icml2021}
\bibliography{main}

\end{document}